\documentclass[sigconf]{acmart}

\usepackage{booktabs}
\usepackage{balance}
\usepackage{array}

\acmConference[SIGSPATIAL '26]{The ACM International Conference on Advances in Geographic Information Systems}{2026}{TBD}
\acmYear{2026}
\copyrightyear{2026}

\title{SeisMamba: Low-Latency Single-Station Seismic Magnitude Estimation for Spatially Distributed Earthquake Early Warning}

\author{Quenton Yeo}
\email{qyeo2651@uni.sydney.edu.au}
\affiliation{%
  \institution{The University of Sydney}
  \country{Sydney, Australia}
}

\author{Zhaoge Bi}
\email{zhaoge.bi@sydney.edu.au}
\affiliation{%
  \institution{The University of Sydney}
  \country{Sydney, Australia}
}

\author{Linghan Huang}
\email{linghan.huang@sydney.edu.au}
\affiliation{%
  \institution{The University of Sydney}
  \country{Sydney, Australia}
}

\author{Luke Stephen Higgins}
\email{luke.higgins@accenture.com}
\affiliation{%
  \institution{Accenture}
  \country{Sydney, Australia}
}

\author{Flora Salim}
\email{flora.salim@unsw.edu.au}
\affiliation{%
  \institution{The University of New South Wales}
  \country{Sydney, Australia}
}

\author{Huaming Chen}
\email{huaming.chen@sydney.edu.au}
\affiliation{%
  \institution{The University of Sydney}
  \country{Sydney, Australia}
}

\begin{document}

\begin{abstract}
Rapid earthquake magnitude estimation is central to earthquake early warning, yet many operational systems depend on dense regional seismic networks and region-specific calibration. This creates a spatial coverage barrier for high-risk areas with sparse sensing infrastructure. Single-station learning offers a lower-cost alternative, but existing models often face an accuracy--latency trade-off and may degrade under regional distribution shift. We present SeisMamba, a lightweight Mamba-based architecture for low-latency magnitude estimation from minimally processed three-component seismic waveforms recorded at a single station. SeisMamba combines hierarchical convolutional encoding, sparse selective state-space modelling, multi-scale feature fusion, and an auxiliary temporal prediction head to support efficient long-sequence waveform analysis. On the STEAD benchmark, SeisMamba achieves the best MSE, RMSE, and $R^2$ among tested baselines while requiring only 0.55 ms for a batch of 32 waveforms on an NVIDIA T4 GPU, making it about three times faster than transformer-based baselines. We further conduct a Chile--Taiwan regional hold-out experiment as a diagnostic test of cross-region deployment, where SeisMamba retains useful performance on geographically unseen seismic regions. These results suggest that selective state-space waveform modelling provides a promising accuracy--latency backbone for spatially distributed, low-cost earthquake early warning.

\end{abstract}

\begin{CCSXML}
<ccs2012>
 <concept>
  <concept_id>10002951.10003260.10003261</concept_id>
  <concept_desc>Information systems~Geographic information systems</concept_desc>
  <concept_significance>500</concept_significance>
 </concept>
 <concept>
  <concept_id>10010147.10010257.10010258.10010259</concept_id>
  <concept_desc>Computing methodologies~Supervised learning</concept_desc>
  <concept_significance>300</concept_significance>
 </concept>
</ccs2012>
\end{CCSXML}

\ccsdesc[500]{Information systems~Geographic information systems}
\ccsdesc[300]{Computing methodologies~Supervised learning}

\keywords{Earthquake, Magnitude, Seismic, Mamba, AI, Spatial}

\maketitle

\section{Introduction}

Earthquake early warning (EEW) aims to estimate event severity quickly enough for alerts to remain useful before damaging shaking reaches exposed locations \cite{cremen_earthquake_2020,wald_practical_2020}. In practice, many EEW systems depend on dense regional seismic networks, historical catalogues, and region-specific calibration, creating a spatial coverage barrier for high-risk regions with sparse monitoring infrastructure. This motivates low-cost single-station learning, where magnitude is estimated from a three-component waveform recorded at one station and the model can run close to the sensing device.

However, single-station EEW is not only a waveform regression problem. It also requires spatially distributed sensing, low-latency local inference, and robustness to regional distribution shift. A model trained on global seismic archives must remain useful when deployed in regions whose waveform characteristics differ from those seen during training. Large datasets such as STEAD provide the scale needed for learning transferable seismic representations \cite{mousavi_stanford_2019}, but spatial variation in attenuation, source properties, site effects, and station conditions still makes cross-region transfer challenging. Existing magnitude-estimation methods address only part of this setting. Traditional and shallow learning approaches often rely on hand-crafted P-wave features or auxiliary source information \cite{zhu_support_2022}. Deep models such as MagNet estimate magnitude directly from raw three-component waveforms \cite{mousavi_machine-learning_2020}, while PhaseNet and EQTransformer provide strong seismic representation backbones originally designed for phase picking or detection \cite{zhu_phasenet_2019,mousavi_earthquake_2020}. More recent attention-based magnitude models improve temporal context modelling \cite{zhang_local_2025}, but recurrent and attention-heavy architectures can increase inference latency, weakening their suitability for edge deployment. Mamba selective state-space models provide a promising alternative by replacing quadratic attention with input-dependent state transitions, enabling linear-time sequence modelling with long-context capacity \cite{gu_mamba_2024}. Recent seismic work has shown that Mamba-style modelling can support phase picking \cite{zhou_phasemamba_2025}. However, its potential for low-latency single-station magnitude estimation under regional deployment shift remains underexplored.

We propose \textit{SeisMamba}, a compact Mamba-based architecture for low-cost single-station earthquake magnitude estimation. Rather than treating magnitude estimation as a standalone waveform regression task, SeisMamba is designed around the requirements of spatially distributed EEW: efficient local inference, strong accuracy--latency balance, and evaluation under regional distribution shift. The model combines convolutional waveform encoding with sparsely placed selective state-space blocks and multi-scale aggregation, while an auxiliary temporal head provides additional supervision on when magnitude evidence becomes stable. Our evaluation studies whether this design can improve the accuracy--latency trade-off on a global seismic benchmark and whether the learned representation remains useful under a geographically held-out regional setting. The results support SeisMamba as an efficient backbone for low-cost, spatially distributed earthquake warning.

\begin{figure*}[ht]
\centering
\includegraphics[width=0.85\linewidth]{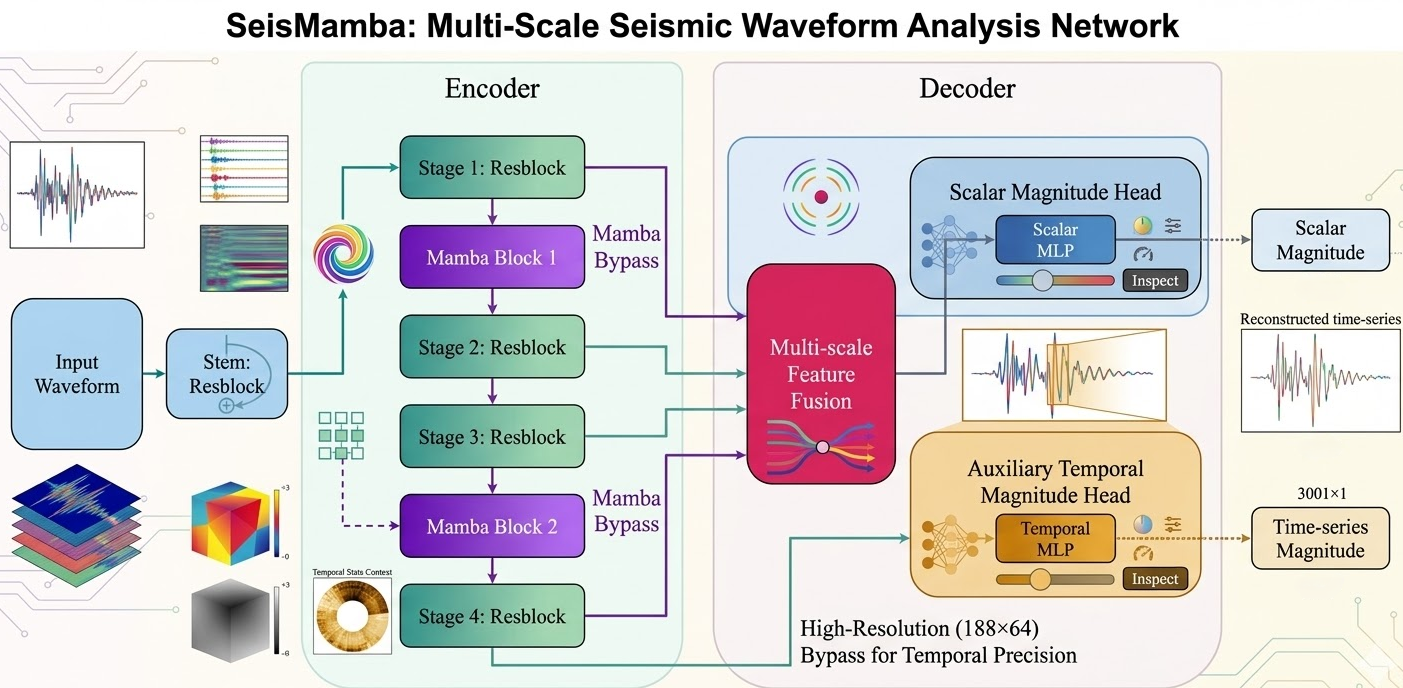}
\caption{SeisMamba architecture overview. Local convolutions extract short-range waveform morphology, sparse Mamba blocks model longer temporal context, and the output heads produce scalar and time-resolved magnitude estimates.}
\Description{A pipeline diagram showing a three-component seismic waveform processed by a convolutional encoder, sparse Mamba blocks, multi-scale feature fusion, and scalar and temporal prediction heads.}
\label{fig:overview}
\end{figure*}

\section{Related Work}

\textbf{Single-station seismic magnitude estimation.}
Single-station EEW reduces reliance on dense regional networks by estimating earthquake properties from local waveform observations. Early learning-based methods include engineered P-wave features, transfer learning, or auxiliary source information to improve rapid magnitude estimation \cite{zhu_support_2022}. MagNet further showed that neural models can estimate magnitude directly from raw waveform data \cite{mousavi_machine-learning_2020}. Related seismic deep models such as PhaseNet and EQTransformer provide strong waveform representation backbones, although they were originally designed for phase picking or detection rather than magnitude regression \cite{zhu_phasenet_2019,mousavi_earthquake_2020}. More recent attention-based models improve temporal context modelling for magnitude estimation \cite{zhang_local_2025}. These studies demonstrate the promise of learning-based EEW, but practical single-station deployment still requires low inference latency and robustness under regional distribution shift.

\textbf{Efficient sequence modelling for spatially distributed seismic sensing.}
Transformers are effective for long-range sequence modelling, but attention has quadratic complexity with sequence length \cite{vaswani_attention_2017}. Mamba provides a selective state-space alternative with linear-time scaling and input-dependent state transitions \cite{gu_mamba_2024}. In seismology, PhaseMamba suggests that Mamba-style models can capture useful waveform structure for phase picking \cite{zhou_phasemamba_2025}. SeisMamba differs from these works by targeting magnitude estimation from single-station waveforms under low-latency and regional deployment constraints, with evaluation on both a global benchmark and geographically held-out seismic regions.

\section{Method}
\subsection{Problem Setup and Design Requirements}
We study single-station earthquake magnitude estimation from minimally processed three-component seismic waveforms. Given a 30-second ENZ waveform $x\in\mathbb{R}^{3\times T}$ sampled at 100 Hz, the model predicts a scalar magnitude $\hat{m}$. Unlike dense-network EEW settings, the target deployment setting assumes that inference may be performed close to a single sensing device. This creates three design requirements: accurate magnitude regression, low inference latency, and robustness when the model is applied to geographically different seismic regions.

We therefore treat the task as both a waveform modelling problem and a regional deployment problem. The standard benchmark measures average predictive performance on STEAD, while the regional hold-out evaluation tests whether the learned representation remains useful when events from selected geographic regions are excluded during training and used only for testing.

\subsection{SeisMamba Architecture}
SeisMamba is designed to balance local waveform feature extraction with efficient long-context sequence modelling. As shown in Fig.~\ref{fig:overview}, it consists of a hierarchical one-dimensional convolutional encoder, sparsely placed Mamba blocks, multi-scale feature fusion, and two prediction heads. The right part of the architecture is a lightweight prediction-side decoder: it maps fused features to a scalar magnitude estimate and an auxiliary time-resolved magnitude trajectory, rather than reconstructing the input waveform.

The encoder first maps the three input channels into a latent feature space. Four residual 1D convolutional stages then progressively reduce temporal resolution while increasing channel width. These stages capture local seismic patterns such as onset sharpness, polarity variation, and amplitude growth. This convolutional front-end is kept lightweight because the target setting requires low-latency local inference.

Mamba blocks are inserted only at selected encoder resolutions. This sparse placement is motivated by efficiency: local convolutions first compress short-range waveform structure, after which selective state-space modelling is applied at shorter sequence lengths. The bypass paths in Fig.~\ref{fig:overview} indicate that not every resolution is processed by a Mamba block; instead, local convolutional features are passed forward directly when long-context modelling is unnecessary. This avoids dense state-space processing at every stage while retaining the ability to model longer waveform context, and it also avoids the quadratic sequence cost of transformer-style attention.

For magnitude prediction, SeisMamba aggregates information across encoder depths. Let $h^{(l)}$ denote the feature map from encoder stage $l$. A pooled descriptor is extracted from each stage and concatenated:
\begin{equation}
z=\mathrm{Concat}\big(\mathrm{Pool}(h^{(1)}),\ldots,\mathrm{Pool}(h^{(4)})\big).
\end{equation}
The scalar head maps $z$ to the final magnitude estimate $\hat{m}$. This multi-scale representation combines shallow onset-sensitive cues with deeper contextual waveform evidence.

An auxiliary temporal head is attached to the deepest feature map, with a high-resolution bypass used to preserve temporal detail for time-resolved prediction. It predicts a magnitude trajectory whose targets are set to zero before the P-wave arrival and to the event magnitude afterwards. This branch provides dense supervision and exposes when magnitude evidence begins to stabilize, but the scalar head remains the primary output used for benchmark comparison. The training objective is
\begin{equation}
\mathcal{L}=0.75\mathcal{L}_{\mathrm{scalar}}+0.25\mathcal{L}_{\mathrm{temporal}},
\end{equation}
where both terms use mean squared error.

\section{Experiments}

\subsection{Dataset, Baselines, and Metrics}
We evaluate SeisMamba on STEAD, a global dataset of single-station three-component seismic waveforms \cite{mousavi_stanford_2019}. Each input is a 30-second ENZ waveform sampled at 100 Hz. To preserve a realistic low-cost deployment setting, we apply only a 1--40 Hz fourth-order zero-phase Butterworth band-pass filter before model input. No hand-crafted P-wave features, source-location inputs, or station-specific calibration are used. All models follow the same preprocessing, random-crop augmentation, labelling procedure, and evaluation protocol. They are trained with AdamW, batch size 32, a base learning rate of $10^{-3}$, a maximum of 150 epochs, early stopping, five warm-up epochs, and a reduce-on-plateau scheduler.

We compare SeisMamba with representative seismic waveform models: MagNet \cite{mousavi_machine-learning_2020}, PhaseNet \cite{zhu_phasenet_2019}, EQTransformer \cite{mousavi_earthquake_2020}, AMAG \cite{zhang_local_2025}, and U-Mamba \cite{ma_u-mamba_2024}. PhaseNet and EQTransformer are adapted to magnitude regression by replacing their original output heads with regression heads. We report MSE, RMSE, MAE, $R^2$, and inference latency. Latency is measured in milliseconds per batch of 32 waveforms on an NVIDIA T4 GPU.
\begin{table}[h]
\centering
\caption{Benchmark comparison on STEAD. Time is measured in ms per batch of 32 waveforms on an NVIDIA T4 GPU. Best values are bold.}
\label{tab:benchmark}
\small
\setlength{\tabcolsep}{3.0pt}
\begin{tabular}{lccccc}
\toprule
Model & MSE & RMSE & MAE & $R^2$ & Time \\
\midrule
PhaseNet & 0.1316 & 0.3627 & 0.2724 & 0.8638 & \textbf{0.50} \\
MagNet & 0.1245 & 0.3528 & 0.2561 & 0.8712 & 0.97 \\
EQTransformer & 0.0860 & 0.2932 & 0.2016 & 0.9237 & 1.70 \\
AMAG & 0.0681 & 0.2609 & \textbf{0.1467} & 0.9389 & 1.58 \\
U-Mamba & 0.0683 & 0.2613 & 0.1793 & 0.9282 & 1.39 \\
SeisMamba & \textbf{0.0628} & \textbf{0.2506} & 0.1566 & \textbf{0.9443} & 0.55 \\
\bottomrule
\end{tabular}
\end{table}
\subsection{Benchmark Accuracy and Latency}
Table~\ref{tab:benchmark} reports the main benchmark results on STEAD. SeisMamba achieves the best MSE, RMSE, and $R^2$ among all tested models, while remaining close to the best MAE. Its inference time is 0.55 ms per batch, only slightly slower than PhaseNet and substantially faster than EQTransformer, AMAG, and U-Mamba. These results support the central claim that sparse selective state-space modelling can improve the accuracy--latency trade-off for single-station magnitude estimation.

\begin{figure}[h]
\centering
\includegraphics[width=0.45\textwidth]{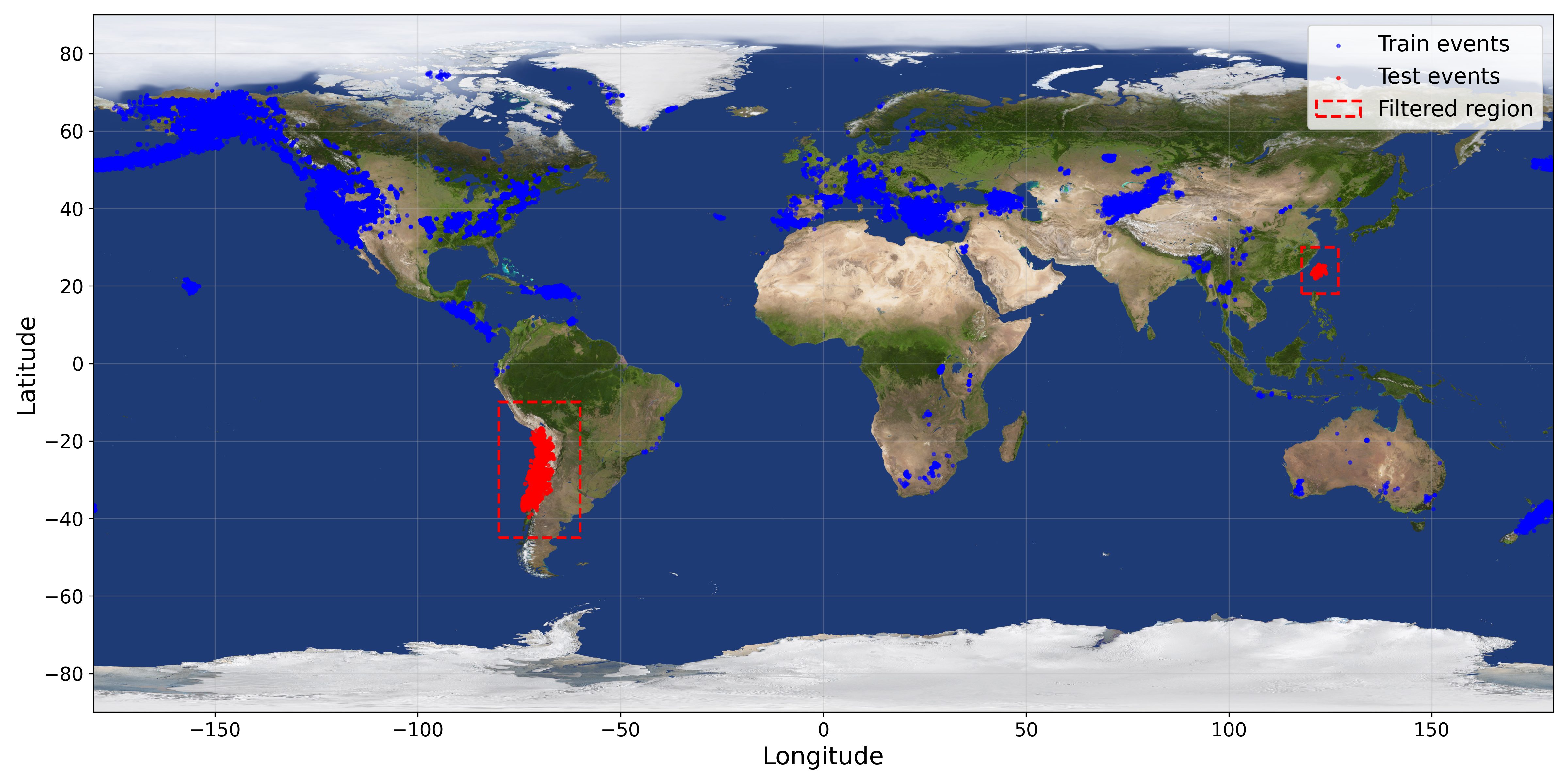}
\caption{Spatial hold-out evaluation setup. Events from Chile and Taiwan are excluded from training and reserved only for geographically unseen testing, while the remaining STEAD events are used for training and validation.}
\Description{World map showing global STEAD training events and held-out test regions in Chile and Taiwan for regional transfer evaluation.}
\label{fig}
\end{figure}

\subsection{Regional Hold-out Evaluation}
Random splits may overestimate deployment readiness because geographically similar events can appear in both training and testing. To probe cross-region deployment more directly, we construct a regional hold-out split by excluding all events from Chile and Taiwan during training and reserving them for testing. This removes 5,478 recordings and creates a geographically unseen evaluation setting. The held-out Chile--Taiwan split contains a broad range of magnitudes, including a substantial proportion of medium-to-large events, making it relevant to practical early-warning scenarios.

As shown in Fig.~\ref{fig}, this split evaluates performance on geographically unseen seismic regions rather than only randomly held-out samples. Table~\ref{tab:transfer_ablation} shows that SeisMamba reaches $R^2=0.8518$ and MAE $=0.2808$ under this harder setting. The performance drop from the standard benchmark is expected because attenuation, site effects, source properties, and station conditions vary across regions. However, the model retains useful predictive ability and its latency remains nearly unchanged at 0.58 ms. This result does not imply full region invariance; rather, it provides a diagnostic stress test showing that the learned representation remains informative beyond the regions seen during training.

\subsection{Ablation Study}
The ablation results in Table~\ref{tab:transfer_ablation} show that the final design is not simply a larger encoder. A deeper variant slightly improves accuracy over the wider variant but nearly doubles inference latency. Dense Mamba insertion performs worse than sparse placement, suggesting that state-space modelling is most useful after local waveform compression. Removing multi-scale fusion also degrades accuracy, confirming that shallow and deep waveform features provide complementary evidence for magnitude estimation.

\begin{table}[h]
\centering
\caption{Regional hold-out and ablation results. Time is measured in ms per batch of 32 waveforms.}
\label{tab:transfer_ablation}
\small
\setlength{\tabcolsep}{3.1pt}
\begin{tabular}{lccccc}
\toprule
Setting & MSE & RMSE & MAE & $R^2$ & Time \\
\midrule
\multicolumn{6}{l}{\textit{Regional hold-out}} \\
Chile--Taiwan & 0.1669 & 0.4085 & 0.2808 & 0.8518 & 0.58 \\
\midrule
\multicolumn{6}{l}{\textit{Ablation on STEAD}} \\
Encoder only & 0.0763 & 0.2762 & 0.1776 & 0.9323 & 0.56 \\
Wider & 0.0669 & 0.2587 & 0.1608 & 0.9406 & \textbf{0.55} \\
Deeper & 0.0660 & 0.2569 & 0.1608 & 0.9414 & 1.09 \\
Dense Mamba & 0.0796 & 0.2822 & 0.1800 & 0.9293 & 0.61 \\
No fusion & 0.0778 & 0.2789 & 0.1798 & 0.9310 & 0.66 \\
Final hybrid & \textbf{0.0628} & \textbf{0.2506} & \textbf{0.1566} & \textbf{0.9443} & \textbf{0.55} \\
\bottomrule
\end{tabular}
\end{table}

\section{Discussion}

The results suggest two main implications for low-cost earthquake early warning. First, selective state-space modelling is a suitable backbone for single-station magnitude estimation because it improves predictive accuracy while preserving low inference latency. This is important for spatially distributed sensing settings, where models may need to run close to individual stations rather than relying entirely on centralized dense-network processing. Second, the regional hold-out experiment provides evidence that SeisMamba learns waveform representations that remain informative beyond random train--test splits. The performance drop under the Chile--Taiwan hold-out also shows that regional waveform shift is not solved: attenuation, site effects, source properties, and station conditions still affect transfer to unseen seismic regions.

The auxiliary temporal head should be interpreted as an observability mechanism rather than a complete uncertainty estimator. It provides time-resolved magnitude supervision and exposes how the model's estimate evolves as waveform evidence accumulates. This is useful for understanding when magnitude information becomes stable, but operational EEW would still require calibrated uncertainty estimates and decision rules before such trajectories could be used for alerting.

% Several limitations remain. SeisMamba estimates magnitude only, whereas a complete EEW system also requires detection, phase picking, localization, alerting, latency analysis, and operational validation. The fixed 30-second window may miss later coda information for large events. Broader cross-region tests, matched baseline comparisons, and limited local fine-tuning are also needed before deployment.

\section{Conclusion}

This paper presented SeisMamba, a lightweight selective state-space architecture for single-station earthquake magnitude estimation. By combining local waveform encoding, sparse Mamba sequence modelling, multi-scale fusion, and auxiliary temporal supervision, SeisMamba improves the accuracy--latency trade-off on a global seismic benchmark while retaining useful performance under a regional hold-out setting. These findings indicate that Mamba-based waveform models are promising components for scalable, low-cost, and spatially distributed earthquake early warning, while also highlighting the need for broader regional adaptation and full-system EEW validation.

\section*{Disclosure of AI Tool Use}
Generative AI tools were used to assist with spelling and grammar checking.

\balance
\bibliographystyle{ACM-Reference-Format}
\bibliography{seismamba_sigspatial2026_short}

\end{document}